\documentclass[journal]{IEEEtran}

\usepackage{cite}

\ifCLASSINFOpdf
  \usepackage[pdftex]{graphicx}
\else
\fi
\usepackage{amsmath}
\usepackage{amsmath, amssymb}
\usepackage{algorithmic}
\usepackage{algorithm}
\usepackage[normalem]{ulem}
\usepackage{array}
\ifCLASSOPTIONcompsoc
  \usepackage[caption=false,font=normalsize,labelfont=sf,textfont=sf]{subfig}
\else
  \usepackage[caption=false,font=footnotesize]{subfig}
\fi
\usepackage{multirow}
\DeclareMathOperator*{\argmax}{arg\,max}
\DeclareMathOperator*{\argmin}{arg\,min}

\newcommand{\tabincell}[2]{\begin{tabular}{@{}#1@{}}#2\end{tabular}}

\begin{document}
\title{Concept Drift Adaptation by Exploiting Historical Knowledge}
\author{Yu~Sun,~\IEEEmembership{Student Member,~IEEE,}
        Ke~Tang,~\IEEEmembership{Senior Member,~IEEE,}
        Zexuan~Zhu,
        and~Xin~Yao,~\IEEEmembership{Fellow,~IEEE}
\IEEEcompsocitemizethanks{\IEEEcompsocthanksitem Y. Sun and K. Tang are with the USTC-Birmingham Joint Research Institute in Intelligent Computation and its Applications, School of Computer Science and Technology, University of Science and Technology of China, Hefei 230027, China. E-mail: sunyu123@mail.ustc.edu.cn; ketang@ustc.edu.cn.\protect\\
\IEEEcompsocthanksitem Z. Zhu is with the College of Computer Science and Software Engineering, Shenzhen University, Shenzhen 518060, China. E-mail: zhuzx@szu.edu.cn.\protect\\
\IEEEcompsocthanksitem X. Yao is with the Centre of Excellence for Research in Computational Intelligence and Applications (CERCIA), School of Computer Science, The University of Birmingham, Edgbaston, Birmingham B15 2TT, UK. E-mail: X.Yao@cs.bham.ac.uk.}
}
\markboth{}%
{}
\maketitle

\begin{abstract}
Incremental learning with concept drift has often been tackled by ensemble methods, where models built in the past can be re-trained to attain new models for the current data. Two design questions need to be addressed in developing ensemble methods for incremental learning with concept drift, i.e., which historical (i.e., previously trained) models should be preserved and how to utilize them. A novel ensemble learning method, namely Diversity and Transfer based Ensemble Learning (DTEL), is proposed in this paper. Given newly arrived data, DTEL uses each preserved historical model as an initial model and further trains it with the new data via transfer learning. Furthermore, DTEL preserves a diverse set of historical models, rather than a set of historical models that are merely accurate in terms of classification accuracy. Empirical studies on 15 synthetic data streams and 4 real-world data streams (all with concept drifts) demonstrate that DTEL can handle concept drift more effectively than 4 other state-of-the-art methods.
\end{abstract}

\IEEEpeerreviewmaketitle

\section{Introduction}
\IEEEPARstart{M}{achine} learning tasks for which training data are available continuously in time have attracted growing attentions due to their wide existence in real-world scenarios, e.g., medical informatics \cite{6137436}, financial data analysis \cite{financial}, social networks \cite{sunyu}, et al. \textit{Incremental learning}, which updates learning machines (models) when a chunk of new training data arrives, is a major learning paradigm for tackling such tasks. In particular, the learning machines should be updated without access to previous data, such that there is no need to store or re-process the previous data \cite{learnpp}, \cite{hehaibo}.

Assuming a number of data chunks $D_1 \cdots D_t$ are available sequentially, the incremental learning procedure is composed of $t$ sub-tasks, each of which can be regarded as a traditional learning task with a distinct data chunk as the training data. Although these sub-tasks can be tackled independently, i.e., training a model (e.g., a classifier) from scratch for each sub-task, it is natural to ask whether the knowledge gained in one sub-task can be leveraged to benefit solving future sub-tasks. Ensemble methods \cite{learnpp}, \cite{sea} offer a natural approach to incremental learning, as each model obtained during the course of incremental learning could be preserved as a base learner and be utilized for solving future sub-tasks. It can be observed from the literature \cite{aue2} that ensemble methods have been used frequently in many advanced incremental learning algorithms and have achieved great successes.

If all the data chunks are generated from the same underlying distribution, ensemble methods for incremental learning are not much different from those for traditional batch learning, i.e., different training data are fed to different base learners. However, the underlying distributions may be non-stationary in real-world applications, since the environment, where data are generated from, may change over time. For example, in the click prediction task of a news website, a breaking news may attract more attention from the visitors and the links of this news are more likely to be clicked. This phenomenon, referred to as \textit{concept drift}, is one of the key challenges that incremental learning approaches \cite{gamasurvey} \cite{zliobaite2}, including those based on ensembles, need to deal with. Specifically, two research questions need to be answered when designing an ensemble method for incremental learning with concept drift, i.e., which historical (i.e., previously trained) models should be preserved for future use and how to exploit the preserved models to facilitate future learning with concept drift.     

To address the above two questions, this paper first reviews the latest progresses on ensemble methods for incremental learning, and then proposes a Diversity and Transfer based Ensemble Learning approach (DTEL). DTEL employs a decision tree as the base learner and a diversity-based strategy for preserving historical models. When a new data chunk arrives, the preserved models are exploited as ``initial points'' for searching/training new models. Finally, the newly obtained models are combined to form the new ensemble. 

The rest of this paper is organized as follows. Section II introduces the notations and formal definitions of the learning problem considered in this paper and reviews related work. The DTEL approach is presented in Section III. Section IV reports our empirical studies on DTEL and other state-of-the-art methods. Section V concludes the paper with directions for future work. 

\section{Related Work}
\subsection{Basic Concepts and Notations}
In incremental learning, at each time step $t$, a chunk of data $ D_t = \{(\mathbf{x}_1^t, y_1^t), (\mathbf{x}_2^t, y_2^t), \cdots, (\mathbf{x}_n^t, y_n^t)\} $, generated from distribution $p_t(\mathbf{x}, y)$, is received, where $\mathbf{x}_i^t$ is a vector of attribute values and $y_i^t$ is a class label. Concept drift can thus be defined as the change of the underlying distribution, e.g., $p_t(\mathbf{x}, y) \neq p_{t-1}(\mathbf{x}, y)$. It should be noted that a special case of a data chunk is a single data example, which is more often referred to as online learning \cite{onlineshuo}. 

At each time step $t$, the learning goal of incremental learning is similar to that of batch learning, i.e., to obtain a good model $F_t$ for $p_t(\mathbf{x}, y)$, which can be stated as
\begin{equation}
F_t = \argmin_{f\in\mathcal{H}} \mathbb{E}_{(\mathbf{x},y)\in p_t(\mathbf{x},y)}[\ell(f(\mathbf{x}),y)],
\label{equ_1}
\end{equation}
where $\mathcal{H}$ is the hypothesis set, $\mathbb{E}(\cdot)$ denotes the expected value of a random variable, and $\ell(\cdot,\cdot)$ is the loss function.
Considering a sequence of $p_t(\mathbf{x}, y)$, the goal of the whole incremental learning process is given by Eq. (\ref{equ_2}):
\begin{equation}
\min_{F_1,F_2,\cdots,F_t,\cdots}\sum_{t} \mathbb{E}_{(\mathbf{x},y)\in p_t(\mathbf{x},y)}[\ell(F_t(\mathbf{x}),y)]
\label{equ_2}
\end{equation}
From Eq. (\ref{equ_2}), it can be observed that at least $t$ models will be generated during the course of incremental learning, which sets a natural basis for ensemble learning.

\subsection{Concept Drift Handling Techniques}
There exist several strategies for handling concept drift in incremental learning, including a sliding window, concept drift detection and ensemble methods. The sliding window methods \cite{hiddencontex}, \cite{vfdt}, \cite{25}, which are mainly applied in the online learning scenario, preserve part of the most recently arrived data and update the current model with both the preserved data and the newly arrived training example. Some other methods \cite{ddm}, \cite{eddm}, \cite{ddd} explicitly involve a concept drift detection module in the learning algorithm. If no concept drift is detected, the current model is updated with newly arrived data. Otherwise, the current model, which may be either a single learner \cite{ddm} or an ensemble \cite{ddd}, is discarded and a new model is built from scratch. 

Neither of the above two strategies requires preserving knowledge/models obtained previously. However, one can easily imagine cases where preserving historical models would be beneficial. For example, if a previously occurred concept appears later (i.e., $p_{t}(\mathbf{x}, y) \neq p_{t-1}(\mathbf{x}, y)$ and $p_{t+1}(\mathbf{x}, y) = p_{t-1}(\mathbf{x}, y)$), a historical model could be used directly. In a more general case, if two concepts are correlated but do not appear consecutively in time, it might be easier to adapt a historical model than further training the current model or building a new one. For such reasons, ensemble methods, which preserve historical models, are gaining more popularity in recent years \cite{sea}, \cite{aue2}, \cite{polikar2}. 

This paper focuses on incremental learning approaches that preserve historical models and exploit them to form ensembles. Typical examples of such approaches include the Streaming Ensemble Algorithm (SEA) \cite{sea}, the Temporal Inductive Transfer (TIX) approach \cite{tixmodel}, the Dynamic Integration of Classifiers (DIC) approach \cite{dynamic}, the Learn++ algorithm \cite{learnpp} in Non-Stationary Environments (Learn\textsuperscript{++}.NSE \cite{polikar2}) and Accuracy Updated Ensemble (AUE2) \cite{aue2}. Although the idea of ensembles is also adopted in the Diversity for Dealing with Drifts (DDD) method \cite{ddd}, we distinguish it from the above-mentioned ensemble methods as DDD does not preserve historical models and the ensembles used in that context could be regarded as a single model for time step $t$.

Due to the concept drift, it is clear that historical models may introduce both positive and negative effects to learning the current concept. Hence, a key issue for all the above ensemble methods is how to get benefits from historical models while preventing negative effects. Suppose a new chunk of data has arrived at time step $t$ and a pool of historical models trained in some previous time steps are available. SEA, DIC, AUE2 and Learn\textsuperscript{++}.NSE exploit the historical model in a similar way. That is, the outputs of the historical models are combined with the output of a new model built using $D_t$ to form final decisions on testing examples of the current concept. These methods differ only in the way that the outputs are combined. 

In SEA, a simple majority voting is utilized. DIC combines the historical models with the new model $D_t$ through Dynamic Selection (DS), Dynamic Voting (DV) or Dynamic Voting with Selection (DVS). For each testing example, its $k$ nearest neighbors in $D_t$ are first identified. Then, the local performance of each individual model is estimated based on these nearest training examples. For each testing example, DS chooses the individual model with the best local performance for classification. DV does not select a single model, but combines the output of all individual models using a weighted voting scheme. DVS is similar to DV, but only takes a half of the individual models that have the best local performance and applies DV to them. AUE2 also employs weighted voting as the combination scheme, where the weights assigned to individual models are determined by mean squared errors of the models. In Learn\textsuperscript{++}.NSE, the weight assigned to an individual model not only depends on its performance on the current training data (as DIC and AUE2), but also depends on its performance on previous data. Specifically, the weight is dynamically adapted as the logarithm of the reciprocal of the model's error on the current and past data chunks, so that the models that performed well on the data that arrive more recently would generally be assigned larger weights when forming an ensemble for the current concept. TIX employs a different method to leverage historical models. That is, given a new chunk of training data $D_t$, the outputs of the historical models on $D_t$ are used as new features of $D_t$, and a new model is built with the augmented $D_t$. If only linear models are built during the learning process, TIX could also be viewed as a special weighted voting scheme, since a linear combination of the original features of $D_t$ can be viewed as the outputs of a linear model on the original $D_t$.

Preserving historical models induces overhead in terms of both storage and computation, e.g., repeatedly assessing the performance of historical models on new training data. Hence, the number of preserved models should be subject to some constraints, instead of increasing unlimitedly. Specifically, given a predefined largest number of preserved models, a selection scheme is needed to decide which historical models should be preserved. This issue is not explicitly addressed in DIC, TIX and Learn\textsuperscript{++}.NSE. Although the DS/DVS scheme of DIC and the time-adjust errors scheme of Learn\textsuperscript{++}.NSE could be adapted to select historical models to preserve, the effectiveness of such adaptations have not be assessed. SEA and AUE2, in contrast, explicitly control the number of preserved models under a predefined threshold. To be specific, both SEA and AUE2 preserve all historical models if the size of the historical model set is smaller than the predefined value. Otherwise, when a new model is trained, a rule, which assesses the quality of both the preserved models and the new model, is employed to decide whether the new model should replace a previously preserved model. Both approaches measure the quality of individual models from the accuracy perspective. The difference is that SEA considers the overall accuracy of ensembles (obtained by replacing a preserved model with the new model) on the current training data, while AUE2 evaluates each individual model under consideration on the current training data directly. None of the existing ensemble methods for incremental learning has considered  ensemble diversity explicitly, although diversity has been shown to play a crucial role in ensembles \cite{yaosurvey}, \cite{tang}.

\section{The Proposed Approach}

The existing ensemble methods for incremental learning, as discussed in Section II, share a common weakness. That is, when a new data chunk arrives, all the approaches utilize the preserved historical models without adapting them to the new training data. The existing approaches differ only in the schemes for combining the outputs of the historical models to fit a new model/ensemble on the current concept. Albeit simple and easy-to-implement, using historical models without any change might not be the best way to exploit these models. Assuming a historical model is established from concept $C_1$, which is quite different from the current concept $C_2$, this historical model may not perform well on the current data, and a good combination scheme will reduce the effect of the historical model on the final ensemble, e.g., assign a small weight to its output. In the extreme case (a weight of 0), the knowledge contained in this historical model is not exploited at all. However, viewing concepts $C_1$ and $C_2$ from the same incremental learning task as the source and target domains of transfer learning\mbox{\cite{transfer}}, it is reasonable to assume that $C_1$ and $C_2$ are correlated with each other. Then, the historical model could still be adapted to the current concept via knowledge transfer. More importantly, an appropriate transfer learning method in this case might even benefit learning concept $C_2$ in terms of accuracy, learning efficiency, or both. Motivated by this consideration, we propose that the historical models are employed as the initial candidate models for building models for new concepts, which is the core idea behind the DTEL approach proposed in this paper.

\begin{figure}[!t]
\centering
\subfloat[Learning flow of the traditional chunk-based ensemble methods.]{
\begin{minipage}[t]{0.5\textwidth}
\centering
\includegraphics[width=2.5in]{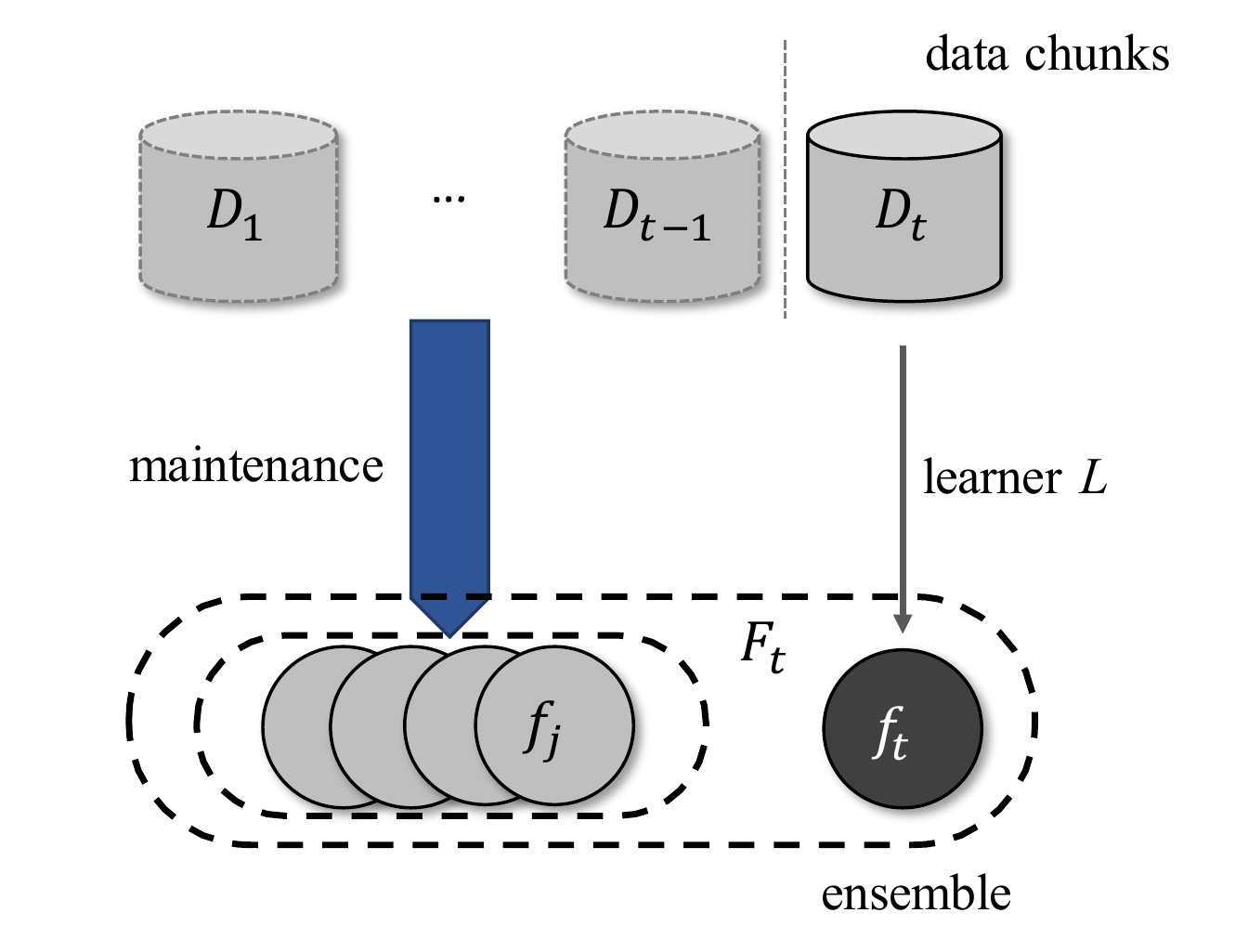}
\end{minipage}
}\\
\subfloat[Learning flow of DTEL.]{
\begin{minipage}[t]{0.5\textwidth}
\centering
\includegraphics[width=2.5in]{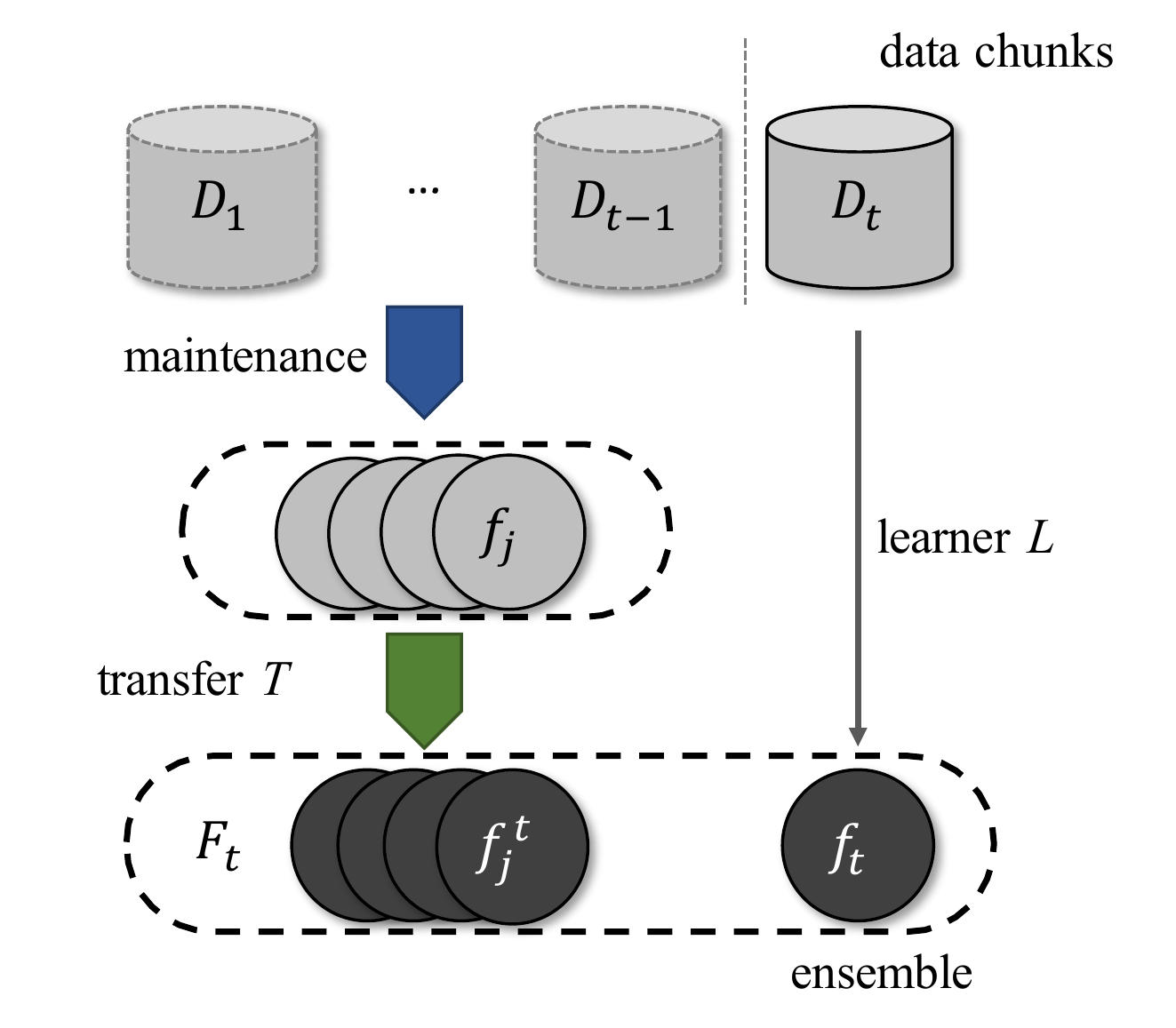}
\end{minipage}
}
\caption{Illustration of the learning flow.} 
\label{framework}
\end{figure}

The framework of DTEL, as given in Algorithm \ref{alg_framework}, differs from the other ensemble methods for incremental learning in two aspects. 
\begin{algorithm}[!t]
\caption{Framework of DTEL}
\begin{algorithmic} [1]
\REQUIRE ~~$(D_1, D_2, \cdots, D_t, \cdots)$: the data stream in incremental learning, $S$: a set of preserved historical models\\
\ENSURE ~~$F_t$, the generalized ensemble model at each time step $t$\\
\WHILE {data chunk $D_t$ is available}
\STATE train a new base model $f_t$ with $D_t$
\STATE obtain the transferred models $f_i^t$ by transferring the preserved historical models $f_i \in S$
\STATE construct the ensemble model $F_t$ with the transferred models $f_i^t$ and the newly trained model $f_t$
\STATE update $S$ with $f_t$ to maximize the diversity and meet the requirement of the archive size
\ENDWHILE
\end{algorithmic}
\label{alg_framework}
\end{algorithm}
First, DTEL does not directly combine the outputs of historical models. Instead, each preserved historical model is first adapted to fit the current data, and then the adapted models and the model constructed from scratch are combined, as illustrated in Fig. \ref{framework}. 
Second, historical models are preserved according to a diversity-based criterion, rather than an accuracy-based criterion. These two key components as well as concrete steps of DTEL are detailed in the following sections.

\subsection{A Diversity-based Model Preservation}

Choosing historical models to preserve in the DTEL framework can be stated as a general problem of selecting $m$ historical models out of $M$, where selected models will be employed as the initial models to be further trained with a new coming data chunk (possibly generated by a drifted concept) and be combined to form an ensemble. In this context, it is the performance of the adapted models after further training that matters, rather than the performance of the original models on the new-coming data chunk. It is well acknowledged in the ensemble learning literature \cite{yaosurvey}, \cite{tang} that, with an appropriate combination scheme, diversity among individual models are essential. Diversity between individual models should be encouraged, which could be implemented by diverse training data, diverse initial models, different learning algorithms \cite{yaosurvey}. In fact, diversity may play an even more important role in DTEL in case of concept drift. Specifically, without prior knowledge regarding the correlations between different concepts, which are nontrivial to know in practice, a historical model that can be nicely adapted to a new data chunk may not be easily adapted to other concepts that may occur in the future. Since the combination scheme functions as a filter to prevent bad individual models deteriorating the performance of an ensemble, a good final ensemble could still be obtained as long as at least one of the preserved historical models could be adapted to the new concept. For this reason, it is desirable that the preserved historical models have sufficient diversity to deal with different concepts, instead of requiring all of them to adapt to the current concept. Therefore, DTEL employs diversity between historical models, instead of the accuracy of either individual models or the ensemble built for the current concept, as a principle for preserving historical models.

Following the diversity-based principle, DTEL preserves historical models using a two-stage strategy. Let historical models be preserved in an archive of size $m$. When a data chunk $D_t$ arrives at time step $t$, the preserved models will be tested on $D_t$ and a new model $f_t$ will be built from scratch using $D_t$. The newly built model $f_t$ will be directly preserved if the archive is not full. Otherwise, $f_t$ will be temporarily incorporated into the archive. Then, the model whose removal will lead to the largest diversity between the remaining models is removed from the archive. In general, any diversity measure \cite{tang} proposed for ensemble learning could be used for this purpose. In this work, the Yules Q-statistic \cite{Yule257} is employed since it is one of the most popular diversity measures in the literature. Concretely, a model (either a historical or the new one) is removed to maximize Eq. (\ref{eq_div}):
\begin{equation}
\text{div}(S) = 1-\frac{1}{\sum_{1\leq i \neq j \leq m}1}\sum_{1\leq i \neq j \leq m}Q(f_i, f_j),
\label{eq_div}
\end{equation}
where $Q(f_i, f_j)$ is the Q-statistic value between $f_i$ and $f_j$, and is calculated by:
\begin{equation}
Q(f_i, f_j) = \frac{N^{11}N^{00}-N^{01}N^{10}}{N^{11}N^{00}+N^{01}N^{10}},
\label{eq_qstat}
\end{equation}
where $N^{ab}$ is the number of examples for which the classification result is $a$ by $f_i$ and $b$ by $f_j$, 1 represents a correct classification and 0 represents a misclassification.

\subsection{Adapting Historical Models through Knowledge Transfer}
Since different learning machines have different learning mechanisms, adapting a historical model to a new data chunk is a model-dependent problem. DTEL employs decision tree as its base learner. A knowledge transfer method has been designed to adapt a previously trained decision tree to a new data chunk. Recall that the process of growing a decision tree incrementally splits the feature space into small regions. Each region corresponds to a leaf node of the tree and is assigned a class label. The structure of the tree inherently contains the knowledge learned from previous sub-tasks, while the class labels assigned to the obtained regions determine the decision boundary. Hence, the proposed knowledge transfer method aims to preserve the structure of a historical decision tree, while perturb it to fit the new data chunk. To be specific, this is done in two steps, as detailed below.

\begin{enumerate}
\item Step 1: Place all of the examples in the new data chunk $D_t$ into the leaf nodes, and reset the class labels of the nodes correspondingly. This step could be viewed as requiring the structure of the adapted decision tree to be as similar to that of the original \cite{Li2008}, such that the knowledge contained in the original tree could be maintained. 
\item Step 2: To meet the requirement of correctly classifying the data in chunk $D_t$, further train a sub-tree in the leaf node in which the stopping criteria has not been reached.
\end{enumerate}

It should be noted that an adapted decision tree derives the knowledge from the current data and the corresponding historical data. Hence, it represents a hybrid knowledge which may not exactly belong to a certain data distribution in the incremental learning task. Besides, the adapted decision trees always fit the current data and are less diverse than the preserved historical trees, since the latter were built with different data chunks. For the reasons given above, the adapted trees will not be preserved, i.e., they will be discarded when the next chunk of data arrives.

\subsection{Detailed Steps of DTEL}

Given the two key components described in Sections III-A and III-B, the detailed steps of DTEL are presented in Algorithm \ref{DTEL}. Suppose $t$ data chunks $D_1 \cdots D_t$ arrive sequentially. The classification and regression tree (CART) \cite{cart}, is employed as the base learner in DTEL. DTEL first builds a decision tree, denoted as $f_1$, with the first chunk of data and preserve $f_1$ in an archive. Then, when a new data chunk, say $D_t$ arrives, the preserved decisions tree(s) is adapted to $D_t$ and a new decision tree $f_t$ is built from scratch based on $D_t$. The adapted trees and the new tree are combined to form the final ensemble for time step $t$. Meanwhile, the new tree $f_t$ is used to update the archive of the preserved historical models. Among the combination schemes described in Section II, the weighted voting scheme used by AUE2 is employed because AUE2 showed the best overall performance among ensemble methods for incremental learning \cite{aue2}. Specifically, the weights for individual adapted trees are determined using Eq. (\ref{eq_weight}). Since the new decision tree $f_t$ is treated as a ``perfect'' classifier, the weight for $f_t$ is calculated according to Eq. (\ref{eq_weightft}).
\begin{equation}
w_{i}^{t}=\frac{1}{\text{MSE}_r^t +\text{MSE}_{i}^{t}+\epsilon},
\label{eq_weight}
\end{equation}
\begin{equation}
w_{t}=\frac{1}{\text{MSE}_r^t+\epsilon},
\label{eq_weightft}
\end{equation}
where $\text{MSE}_{i}^{t}$ estimates the prediction error of the adapted tree $f_i$ on data chunk $D_t$, $\text{MSE}_{r}^t$ is the mean square error of a random classifier, and $\epsilon$ is a very small positive value to avoid the denominator of Eq. (\ref{eq_weight}) being 0. $\text{MSE}_{i}^{t}$ and $\text{MSE}_{r}^t$ are calculated as follows in Eq. (\ref{eq_mseij}) and (\ref{eq_mser}),
\begin{equation}
\text{MSE}_{i}^{t} = \frac{1}{|D_t|}\sum_{\{\textbf{x},y\}\in D_t}(1-p_i^t(y|\textbf{x}))^2,
\label{eq_mseij}
\end{equation}
\begin{equation}
\text{MSE}_r^t = \sum_{y}p^t(y)(1-p^t(y))^2,
\label{eq_mser}
\end{equation}
where $p_i^t(y|\textbf{x})$ denotes the posterior probability given by adapted classifier of $f_i$ on data in chunk $D_t$ and $p^t(y)$ denotes the prior probability of class $y$ in $D_t$. The posterior probability in the tree model is calculated as the ratio of that class in the corresponding leaf node. When the archive of historical trees is full, lines 9-11 will be activated to decide whether a newly trained tree will replace a preserved tree in the archive. 

\begin{algorithm}[!t]
\caption{DTEL}
\begin{algorithmic} [1]
\REQUIRE ~~$(D_1, D_2, \cdots, D_t, \cdots)$: the data stream, $T$: model transfer function, $S_t$: the historical model set at time step $t$, $E_t$: the ensemble model set at time step $t$, $m$: the archive size of the historical model set, $\text{div}$: model maintenance function with diversity\\
\ENSURE ~~$F_t$, the generalized model at each time step $t$\\
\WHILE {data chunk $D_t$ is available}
\STATE $f_t \gets$ train a new base model with $D_t$
\STATE $f_i^t \gets T(f_i, D_t)$, for all $f_i \in S_t$
\IF {$|S_{t-1}|<m$}
\STATE $S_t \gets S_{t-1} \bigcup \{f_t\}$
\ELSE
\STATE $S_t^{'} \gets S_{t-1} \bigcup \{f_t\}$
\STATE $f_{replace} \gets \argmax_{f_i}\text{div}(S_t^{'}- \{f_i\})$
\STATE $S_t \gets S_{t}^{'} - \{f_{replace}\}$
\ENDIF
\STATE $w_i^t \gets$ estimate the weight for each $f_i^t$ by Eq. (\ref{eq_weight})
\STATE $w_t \gets$ estimate the weight for $f_t$ by Eq. (\ref{eq_weightft})
\STATE $F_t = (\sum_{i} w_i^t f_i^t + w_tf_t) / (\sum_{i}w_i^t + w_t)$
\ENDWHILE
\end{algorithmic}
\label{DTEL}
\end{algorithm}

\section{Experiment}
Empirical studies have been conducted to assess the performance of DTEL. Different types of concept drift are involved, and state-of-the-art algorithms are compared in the experiment. The algorithms are evaluated mainly from three aspects, i.e., the classification performance for each chunk, the overall performance for a whole data stream, and the time efficiency.

The following representative algorithms are compared in the experiments: SEA \cite{sea}, Learn\textsuperscript{++}.NSE \cite{polikar2}, AUE2 \cite{aue2}, and TIX \cite{tixmodel}. These approaches employ different methods for concept drift adaptation. SEA, Learn\textsuperscript{++}.NSE and AUE2 are ensemble methods to exploit the historical knowledge, while TIX uses the historical knowledge by introducing it as new features in a transfer manner. SEA and Learn\textsuperscript{++}.NSE use the historical model directly in an ensemble, while AUE2 updates each historical model with the new chunk of data. In addition, for the weight assignment in the ensemble method, SEA uses the uniform weight, AUE2 employs the performance-based weight, and Learn\textsuperscript{++}.NSE uses the time-adjusted performance-based weight.

All of the compared approaches are frameworks. To make the comparisons fair, the decision tree was employed as the base learner in all of these approaches. Specifically, the traditional decision tree method CART \cite{cart} is applied in SEA, Learn\textsuperscript{++}.NSE, TIX, and DTEL. Since AUE2 needs to use an on-line model as the base learner, Hoeffding tree \cite{hoeffding}, an on-line decision tree method, was applied.

In SEA, AUE2, and DTEL, a limited number of historical models are preserved. The archive size of the historical model set is the only parameter to set in the experiment. According to the suggestion in \cite{sea}, the ensemble size is set to 25 for the compared algorithms, unless mentioned otherwise.

\subsection{Comparison on Synthetic Data}
\subsubsection{Datasets}
In order to comprehensively investigate the performance of DTEL, five types of concept drift were tested in the experiment. When working with synthetic data, it is known exactly what the type of concept drift is and how dramatic the data distribution changes. Hence, it is important to use the synthetic data for a detailed analysis of the approaches in concept drift adaptation. Based on the previous research \cite{sea}, \cite{aue2}, \cite{ddm}, \cite{polikar2}, \cite{diversity}, \cite{ref1}, five widely used synthetic concept drifts are employed in our experiment, described as follows.

\begin{table*}[!htbp]
\renewcommand{\arraystretch}{1}
\caption{Artificial Concept Drift.}
\centering
\begin{tabular}{c | c | c}
\hline
Drift Type & Fixed Value & Drift Value\\
\hline
\multirow{2}{*}{SEA} & \multirow{2}{*}{$a=1$, $b=1$} & $\theta = 10 \to 7 \to 3 \to 7 \to 10 \to 13 \to 16 \to 13$\\
\cline{3-3}
&& $\theta = 10 \to 8 \to 6 \to 8 \to 10 \to 12 \to 14 \to 12$\\
\hline
\multirow{2}{*}{ROT} & \multirow{2}{*}{$a=0$, $b=0$} & $\theta=0$, $\Delta \theta = \pi/30$\\
\cline{3-3}
&& $\theta=0$, $\Delta \theta = \pi/60$\\
\hline
\multirow{2}{*}{CIR} & \multirow{2}{*}{$a=0$, $b=0$} & $\theta = 3 \to 2 \to 1 \to 2 \to 3 \to 4 \to 5 \to 4$\\
\cline{3-3}
&& $\theta = 3 \to 2.5 \to 2 \to 2.5 \to 3 \to 3.5 \to 4 \to 3.5$\\
\hline
\multirow{2}{*}{SIN} & \multirow{2}{*}{$a=1$, $b=1$, $c=0$ }& $\theta=0$, $\Delta \theta = \pi/30$\\
\cline{3-3}
&& $\theta=0$, $\Delta \theta = \pi/60$\\
\hline
\multirow{2}{*}{STA} & \multirow{2}{*}{\tabincell{c}{$c=S\land M\land L$\\$\land_2$\\$=_1$, $=_2$, $=_3$}} & \tabincell{c}{$(a=R,\land_1,b=C) \to (a=B,\lor_1,b=C) \to (a=G,\lor_1,b=S) \to$\\$ (a=G,\land_1,b=T) \to (a=G,\lor_1,b=C) \to (a=R,\lor_1,b=S)$}\\
\cline{3-3}
&& \tabincell{c}{$(a=R,\land_1,b=C) \to (a=B,\land_1,b=C) \to (a=B,\lor_1,b=C) \to$\\$ (a=B,\lor_1,b=S) \to (a=B,\land_1,b=S) \to (a=G,\land_1,b=S)$}\\
\hline
\end{tabular}
\label{table_concept_drift}
\end{table*}

\begin{itemize}
\item \textbf{SEA moving hyperplane concepts (SEA)} \cite{sea} involves 3 features with a value between 0 and 10. Only two features (i.e., $x_1$ and $x_2$) are relevant, and $x_3$ is a noisy feature with a random value. The class label of data in this concept is determined by
\begin{displaymath}
ax_1+bx_2 \leq/> \theta
\end{displaymath}
To simulate the concept drift, the value of $\theta$ changes during the learning process.
\item \textbf{Rotating concepts (ROT)} \cite{aue2}, \cite{polikar2} rotates the decision boundary or data points to simulate the change of data distribution. The formulation of rotating the data point in the 2-dimensional feature space around $(a,b)$ is shown as follow.
\begin{displaymath}
x_1 \gets (x_1-a)\cos\theta-(x_2-b)\sin\theta+a
\end{displaymath}
\begin{displaymath}
x_2 \gets (x_1-a)\cos\theta+(x_2-b)\sin\theta+b
\end{displaymath}
In the experiment, a data set with 6 classes is used as the data source, and the rotation is implemented evenly in the learning process.
\item \textbf{Circle concepts (CIR)} \cite{ddm}, \cite{diversity} applies a circle as the decision boundary in a 2-dimensional feature space and simulates the concept drift by changing the radius of the circle, i.e.,
\begin{displaymath}
(x_1-a)^2+(x_2-b)^2 \leq/> \theta
\end{displaymath}
In the experiment, data points are generated evenly locating between -5 and 5 for both dimensions, and the radius value of $\theta$ changes every 25 data chunks.
\item \textbf{Sine concepts (SIN)} \cite{ddm}, \cite{diversity} determines the label of data by a sine curve in a 2-dimensional feature space, which is defined as follow.
\begin{displaymath}
a\sin(bx_1+\theta)+c \leq/> x_2
\end{displaymath}
In the experiment, all of the data locate in the area of [-5, 5] for both dimensions. The value of $\theta$ is evenly changed to generate the change of the data distribution.
\item \textbf{STAGGER Boolean concepts (STA)} \cite{diversity}, \cite{ref1} generates the data with categorical features using a set of rules to determine the class label. According to \cite{diversity} and \cite{ref1}, the features and values are color $\in$ \{red(R), blue(B), green(G)\}, shape $\in$ \{circle(C), square(S), triangle(T)\}, and size $\in$ \{small(S), medium(M), large(L)\}. The decision rules can be formulated as follow.
\begin{displaymath}
\begin{aligned}
y=&((\text{color}=_1 / \neq_1 a)\lor_1 / \land_1(\text{shape}=_2 /\neq_2 b))\\
&\lor_2/\land_2(\text{size}=_3 / \neq_3 c)
\end{aligned}
\end{displaymath}
The concept drift is simulated by changing the items in the rules.
\end{itemize}

The dramatic degree of concept drift and the size of data chunks may influence the performance of the learning algorithm. In this regard, three data streams are generated for each type of concept drift, with different dramatic degrees of concept drift and different chunk sizes. All of the synthetic data streams are generated with 120 data chunks with 10\% noise introduced. The specific setting of the values for simulating the concept drifts are illustrated in Table \ref{table_concept_drift}. The statistics of the synthetic data streams are described in Table \ref{table_syndata}. For synthetic data streams, two chunks of data are generated at each learning step. The first data chunk is used for training and the other one is used to test the current prediction model.

\begin{table}[!t]
\renewcommand{\arraystretch}{1}
\caption{Synthetic Data Streams in Experiment.}
\centering
\begin{tabular}{c c c c c c}
\hline
\multicolumn{1}{c}{Data Stream} & \multicolumn{1}{c}{\#Example} & \multicolumn{1}{c}{\#Feature} & \multicolumn{1}{c}{\#Label} & \multicolumn{1}{c}{Chunk Size}\\
\hline
SEA200A & 24,000 & 3 & 2 & 200\\
SEA200G & 24,000 & 3 & 2 & 200\\
SEA500G & 60,000 & 3 & 2 & 500\\
ROT200A & 24,000 & 2 & 6 & 200\\
ROT200G & 24,000 & 2 & 6 & 200\\
ROT500G & 60,000 & 2 & 6 & 500\\
CIR200A & 24,000 & 3 & 2 & 200\\
CIR200G & 24,000 & 3 & 2 & 200\\
CIR500G & 60,000 & 3 & 2 & 500\\
SIN200A & 24,000 & 2 & 2 & 200\\
SIN200G & 24,000 & 2 & 2 & 200\\
SIN500G & 60,000 & 2 & 2 & 500\\
STA200A & 24,000 & 3 & 2 & 200\\
STA200G & 24,000 & 3 & 2 & 200\\
STA500G & 60,000 & 3 & 2 & 500\\
\hline
\end{tabular}
\label{table_syndata}
\end{table}

\subsubsection{Results}
\begin{figure*}[!t]
\centering
\includegraphics[width=7in]{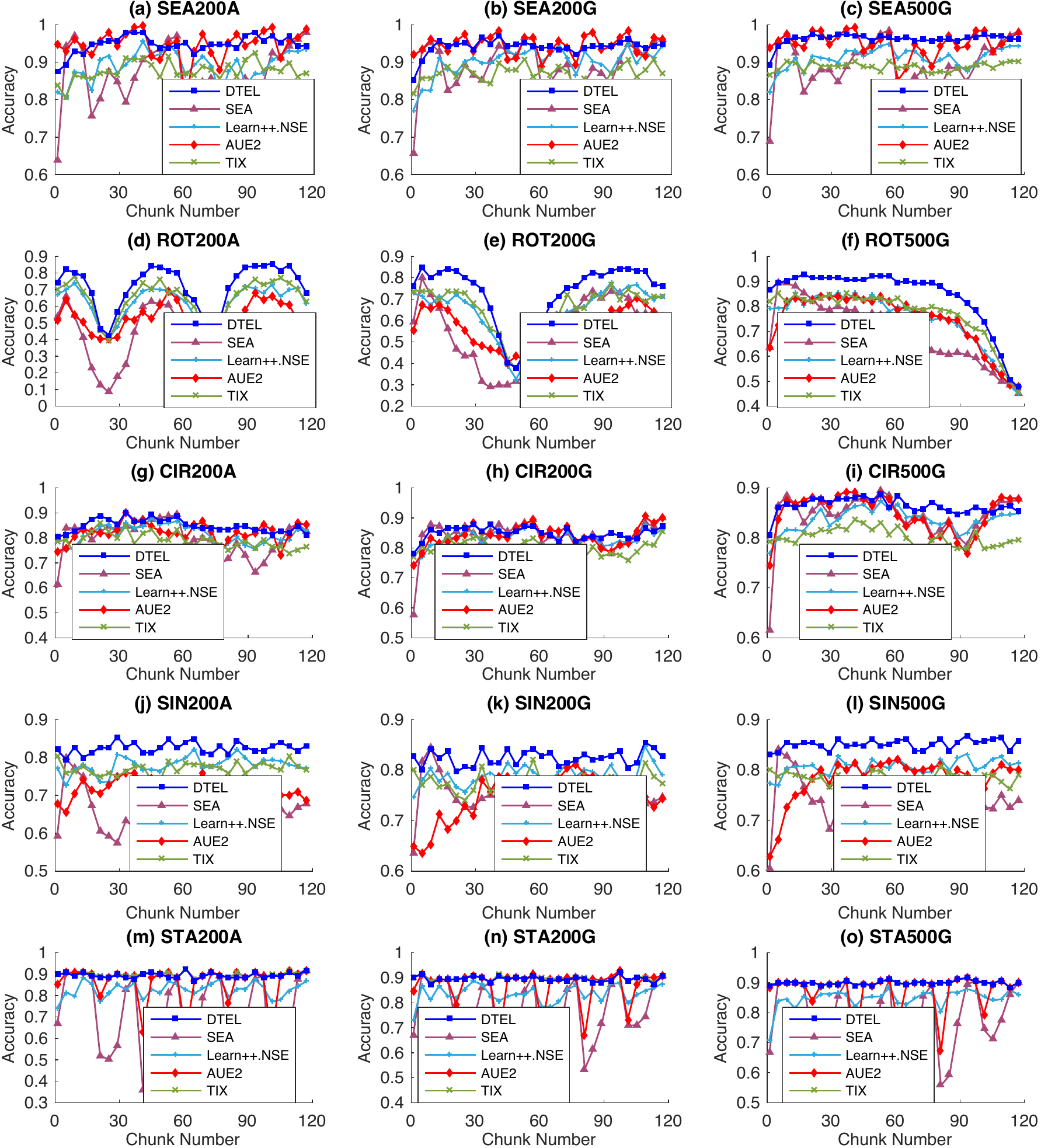}
\caption{Accuracy results on synthetic data streams.}
\label{fig_syndata}
\end{figure*}

In order to investigate the performance of the algorithms in concept drift adaptation, the prediction accuracy in each chunk of data are evaluated and presented in Fig. \ref{fig_syndata}. Generally speaking, the accuracy result obtained from the proposed DTEL is not only the highest among the compared algorithms but also relatively stable across different synthetic data streams. Specifically, for SEA data streams (as shown in Fig. \ref{fig_syndata} ($a$)-($c$)), DTEL performed steadily on each data chunk and was not affected by concept drifts, no matter how dramatic the concept drift is. Although AUE2 obtained the highest accuracy on several of the data chunks, it can be observed that AUE2 is sensitive to the concept drift with a drop in accuracy. All of the compared algorithms are sensitive to the ROT concept drift, with a dramatic fluctuation of classification performance, as shown in Fig. \ref{fig_syndata} ($d$)-($f$). However, DTEL still obtained the highest accuracy on all the data chunks, which illustrates the rapid concept drift adaptation ability of DTEL. It is interesting to note that although the ROT concept drift is generated by smoothly rotating the decision boundary, the performance of the compared algorithms periodically rises and falls, instead of decaying gradually. For the CIR concept drift (Fig. \ref{fig_syndata} ($g$)-($i$)), the performance of the compared algorithms are relatively similar. The dramatic degree of CIR concept drift appears not to influence the performance of DTEL based on the observation of results on CIR200A and CIR200G. In contrast, the other compared algorithms are affected by this type of concept drift. The results obtained from the SIN data streams (as shown in Fig. \ref{fig_syndata} ($j$)-($l$)) also demonstrate the superiority of DTEL, which performed the best on almost all the data chunks. Different from the previously analysed data streams, STA data streams are generated from decision rules. For the results obtained from STA data streams (Fig. \ref{fig_syndata} ($m$)-($o$)), the compared algorithms can be divided into two categories. DTEL and TIX show a high accuracy with a low variance among different data chunks, while the accuracy results of SEA, AUE2 and Learn\textsuperscript{++}.NSE fluctuate dramatically and are more sensitive to this type of concept drift.

To quantitatively assess the performance of the compared algorithms on the whole data stream, the classification accuracy of each algorithm is averaged over the data chunks in a data stream and presented in Table \ref{table_accuracy_syn}. The standard deviations over data chunks, which indicates how stably an algorithm performs on a data stream,  are also given. Generally, DTEL shows a clear advantage over other algorithms, in terms of high accuracy and low standard deviation on most data streams. On SEA200G, AUE2 obtained the highest accuracy, but its standard variance is higher than DTEL, which means AUE2 is more sensitive to concept drift on SEA200G than DTEL. TIX model performed the best on the STA data streams. The reason might be that the data in STA streams are generated by decision rules and the historical knowledge is represented as a new feature value in TIX. By employing decision tree as the learner, TIX can exploit the useful historical knowledge more easily with the tree structure. The Wilcoxon rank-sum test at the 95\% confidence level has been conducted to check whether the differences (in terms of average accuracy) between DTEL and the compared algorithms are statistically significant, and the result is shown in the last row of Table \ref{table_accuracy_syn} as the statistics of win-tie-loss. As shown in the statistical result, DTEL performed statistically significantly better than other algorithms on the synthetic data streams in pairwise comparisons. The average accuracy results in Table \ref{table_accuracy_syn} further verify the detailed observation results of the performance of DTEL on each data chunk shown in Fig. \ref{fig_syndata}. It can be concluded that DTEL is able to handle better the concept drift of different types and different dramatic degrees, by using the historical knowledge.

\begin{table*}[!t]
\renewcommand{\arraystretch}{1}
\caption{Average Accuracy (\%) of Every Chunk ($\pm$ The Standard Deviation of The Accuracy for Each Chunk) on Synthetic Data Streams. $\bullet$/$\circ$ Indicates That DTEL Is Significantly Better/Worse Than The Corresponding Algorithm (Wilcoxon rank-sum test at 95 percent confidence level). The Values in Boldface Indicate the Highest Accuracy on The Corresponding Data Stream.}
\label{table_accuracy_syn}
\centering
\begin{tabular}{c  c  c  c  c  c}
\hline
\multicolumn{1}{c}{Data} & \multicolumn{1}{c}{DTEL} & \multicolumn{1}{c}{SEA} & \multicolumn{1}{c}{Learn\textsuperscript{++}.NSE} & \multicolumn{1}{c}{AUE2} & \multicolumn{1}{c}{TIX}\\
\hline
SEA200A & \textbf{94.77 $\pm$ 2.99} & 86.31 $\pm$ 11.43$\bullet$ & 89.07 $\pm$ 5.13$\bullet$ & 94.66 $\pm$ 4.94 & 87.77 $\pm$ 3.97$\bullet$\\
SEA200G & 94.07 $\pm$ 2.69 & 88.90 $\pm$ 10.02$\bullet$ & 90.02 $\pm$ 4.98$\bullet$ & \textbf{94.58 $\pm$ 3.80}$\circ$ & 86.90 $\pm$ 4.26$\bullet$\\
SEA500G & \textbf{96.21 $\pm$ 1.76} & 89.37 $\pm$ 10.17$\bullet$ & 91.10 $\pm$ 3.45$\bullet$ & 95.02 $\pm$ 4.05 & 88.85 $\pm$ 2.54$\bullet$\\
ROT200A & \textbf{71.86 $\pm$ 13.91} & 37.88 $\pm$ 18.17$\bullet$ & 62.19 $\pm$ 11.49$\bullet$ & 52.72 $\pm$ 9.99$\bullet$ & 65.02 $\pm$ 11.45$\bullet$\\
ROT200G & \textbf{72.67 $\pm$ 14.28} & 54.61 $\pm$ 17.45$\bullet$ & 63.41 $\pm$ 12.84$\bullet$ & 55.43 $\pm$ 9.76$\bullet$ & 64.97 $\pm$ 12.16$\bullet$\\
ROT500G & \textbf{84.27 $\pm$ 12.20} & 69.81 $\pm$ 14.29$\bullet$ & 74.77 $\pm$ 11.57$\bullet$ & 74.34 $\pm$ 11.34$\bullet$ & 76.98 $\pm$ 10.44$\bullet$\\
CIR200A & \textbf{84.84 $\pm$ 3.88} & 79.90 $\pm$ 10.10$\bullet$ & 81.33 $\pm$ 5.79$\bullet$ & 82.21 $\pm$ 5.29$\bullet$ & 78.98 $\pm$ 5.30$\bullet$\\
CIR200G & \textbf{84.88 $\pm$ 3.37} & 83.86 $\pm$ 8.60 & 83.27 $\pm$ 4.38$\bullet$ & 84.06 $\pm$ 4.63 & 80.04 $\pm$ 4.65$\bullet$\\
CIR500G & \textbf{86.42 $\pm$ 2.22} & 84.32 $\pm$ 8.43$\bullet$ & 83.60 $\pm$ 3.00$\bullet$ & 84.87 $\pm$ 4.36$\bullet$ & 80.20 $\pm$ 2.70$\bullet$\\
SIN200A & \textbf{82.51 $\pm$ 2.84} & 65.78 $\pm$ 8.44$\bullet$ & 78.02 $\pm$ 4.00$\bullet$ & 71.91 $\pm$ 3.69$\bullet$ & 77.05 $\pm$ 3.91$\bullet$\\
SIN200G & \textbf{82.40 $\pm$ 3.22} & 74.12 $\pm$ 8.11$\bullet$ & 79.07 $\pm$ 3.27$\bullet$ & 74.32 $\pm$ 5.48$\bullet$ & 77.23 $\pm$ 3.70$\bullet$\\
SIN500G & \textbf{85.10 $\pm$ 1.91} & 73.76 $\pm$ 7.80$\bullet$ & 80.67 $\pm$ 2.51$\bullet$ & 78.22 $\pm$ 4.71$\bullet$ & 78.36 $\pm$ 2.84$\bullet$\\
STA200A & 89.48 $\pm$ 2.34 & 70.07 $\pm$ 21.57$\bullet$ & 82.74 $\pm$ 7.88$\bullet$ & 86.06 $\pm$ 10.93 & \textbf{89.85 $\pm$ 2.01}\\
STA200G & 89.56 $\pm$ 2.50 & 76.01 $\pm$ 15.71$\bullet$ & 83.56 $\pm$ 7.54$\bullet$ & 86.58 $\pm$ 9.06 & \textbf{89.77 $\pm$ 2.32}\\
STA500G & \textbf{90.00 $\pm$ 1.34} & 76.43 $\pm$ 15.35$\bullet$ & 84.78 $\pm$ 7.25$\bullet$ & 87.47 $\pm$ 7.73 & \textbf{90.00 $\pm$ 1.34}\\
\hline
\multicolumn{2}{c}{win-tie-loss} & 14-1-0 & 15-0-0 & 8-6-1 & 12-3-0\\
\hline
\end{tabular}
\end{table*}

\subsection{Comparison on Real-world Data}
\subsubsection{Datasets}
In addition to the synthetic data streams, 4 real-world data streams, namely covertype, poker hand, electricity, and click-through rate prediction data were also employed in our experiments. Details of these datasets are described as follow.

\begin{itemize}
\item \textbf{Covertype} \cite{uci} is a real-world dataset for describing the observation of a forest area with 51 cartographic variables. Six class labels are involved to represent the corresponding forest cover type.
\item \textbf{PokerHand} \cite{uci} describes the suits and ranks of a hand of five playing cards. It involves 5 numerical features and 5 ordinal features for each example. Ten class labels exist in the dataset for describing different poker hands.
\item \textbf{Electricity}, a widely used dataset \cite{ddm}, \cite{aue2}, is collected from the New South Wales Electricity Market in Australia, containing 45,312 instances dated from 7 May 1996 to 5 December 1998. Each example is described by 8 features, and the class label identifies the change of the price (i.e., up and down).
\item \textbf{Click-through Rate Prediction (CTRPrediction)} is a dataset obtained from the Tencent company. All of the examples in the data set are in their original order collected through 30 days. After the pre-processing of the raw data, 20,000 examples are selected for each day, and totally 600,000 examples with 100 features are tested.
\end{itemize}

The chunk size was set to 1,000 for the first three real-world data streams. The chunk size for CTRPrediction data was set to 10,000 and each chunk of data represents half a day's observations from the real-world application. The statistics of the real-world data streams are described in Table \ref{table_data_real}. For real-world data streams, each chunk of data is first used to test the current prediction model and then to update the model in learning. Considering the chunk size is relatively large and the number of chunks in CTRPrediction data is relatively small, the ensemble size was set to 3 for all compared algorithms on this dataset. 

\begin{table}[!t]
\renewcommand{\arraystretch}{1}
\caption{Real-world Data Streams in Experiment.}
\label{table_data_real}
\centering
\begin{tabular}{p{42pt}<{\centering} p{25pt}<{\centering} p{24pt}<{\centering} p{24pt}<{\centering} p{24pt}<{\centering} p{38pt}<{\centering}}
\hline
Data Stream & \#Example & \#Feature & \#Label & \#Chunk & Chunk Size\\
\hline
Covertype & 581,000 & 51 & 7 & 581 & 1,000\\
PokerHand & 1,000,000 & 10 & 10 & 1,000 & 1,000\\
Electricity & 44,000 & 8 & 2& 44 & 1,000\\
CTRPrediction & 600,000 & 100 & 2& 60 & 10,000\\
\hline
\end{tabular}
\end{table}

\subsubsection{Results}
\begin{table*}[!htbp]
\renewcommand{\arraystretch}{1}
\caption{Average Accuracy (\%) of Every Chunk ($\pm$ The Standard Deviation of The Accuracy for Each Chunk) on Real-world Data Streams. $\bullet$/$\circ$ Indicates That DTEL Is Significantly Better/Worse Than The Corresponding Algorithm (Wilcoxon rank-sum test at 95 percent confidence level). The Values in Boldface Indicate the Highest Accuracy on The Corresponding Data Stream.}
\centering
\begin{tabular}{c  c  c  c  c  c}
\hline
\multicolumn{1}{c}{Data} & \multicolumn{1}{c}{DTEL} & \multicolumn{1}{c}{SEA} & \multicolumn{1}{c}{Learn\textsuperscript{++}.NSE} & \multicolumn{1}{c}{AUE2} & \multicolumn{1}{c}{TIX}\\
\hline
Covertype & \textbf{91.80 $\pm$ 8.42} & 71.46 $\pm$ 15.14$\bullet$ & 84.11 $\pm$ 12.45$\bullet$ & 87.09 $\pm$ 8.74$\bullet$ & 88.32 $\pm$ 9.15$\bullet$\\
PokerHand & 51.20 $\pm$ 1.81 & \textbf{56.36 $\pm$ 2.54}$\circ$ & 45.86 $\pm$ 1.87$\bullet$ & 51.31 $\pm$ 1.79 & 47.23 $\pm$ 1.73$\bullet$\\
Electricity & 75.05 $\pm$ 8.36 & 72.35 $\pm$ 13.99 & 75.54 $\pm$ 8.09 & \textbf{76.47 $\pm$ 8.70} & 73.55 $\pm$ 8.72\\
CTRPrediction & \textbf{82.02 $\pm$ 20.90} & 66.08 $\pm$ 19.36$\bullet$ & 77.13 $\pm$ 20.96$\bullet$ & 80.57 $\pm$ 21.34$\bullet$ & 80.57 $\pm$ 21.58$\bullet$\\
\hline
\multicolumn{2}{c}{win-tie-loss} & 2-1-1 & 3-1-0 & 2-2-0 & 3-1-0\\
\hline
\end{tabular}
\label{table_accuracy_real}
\end{table*}

\begin{figure*}[!t]
\centering
\includegraphics[width=7in]{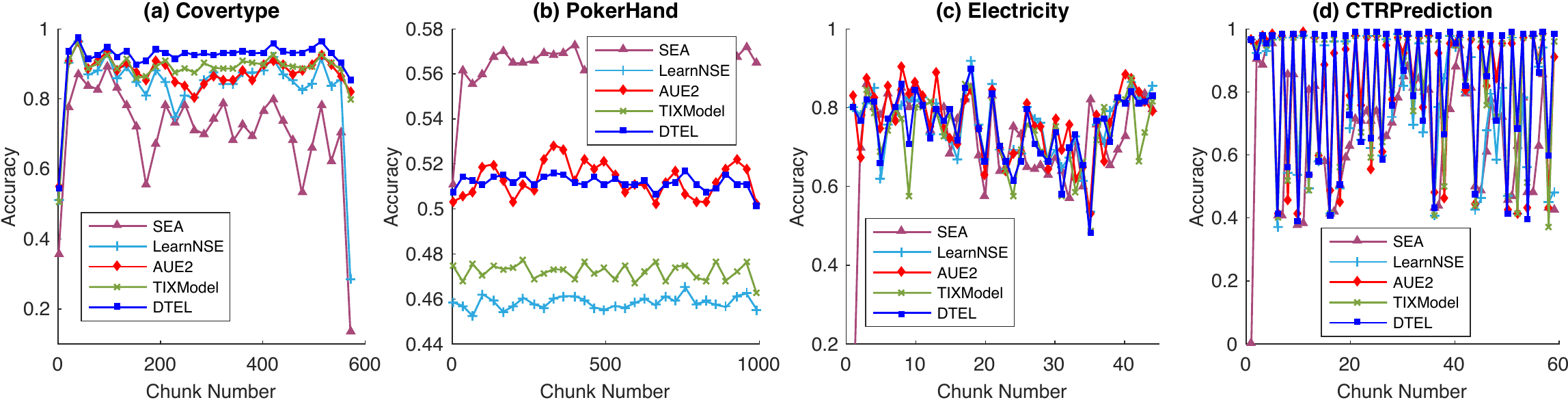}
\caption{Accuracy results on real-world data streams.}
\label{realdata}
\end{figure*}

The accuracy of the compared algorithms is shown in Fig. \ref{realdata}. Since no detailed information regarding the occurrences and behaviors of concept drift is known in the real-world data streams, it is hard to conduct an exact analysis of the adaptation ability for concept drift. Nevertheless, the empirical results on real-world data streams validate the conclusion drawn from the synthetic data. For the Covertype data stream (Fig. \ref{realdata} ($a$)), DTEL showed a stable performance along the whole incremental learning process, while the compared algorithms showed an unstable classification performance, especially the SEA approach. On PokerHand data stream (Fig. \ref{realdata} ($b$)), SEA performs better than DTEL. A possible reason is that random events may happen in poker card game and lead the data distribution in PokerHand data to be almost randomly changed, which is close to the assumption in SEA approach. On PokerHand data, DTEL generally performed more stable than AUE2 and obtained a better performance on all of the data chunks than Learn\textsuperscript{++}.NSE and TIX. The data streams of Electricity are relatively hard to learn, on which the compared algorithms performed similarly and unstably. It is hard to distinguish the best performed algorithm in Fig. \ref{realdata} ($c$). On CTRPrediction data (Fig. \ref{realdata} ($d$)), the performance of all of the compared algorithms fluctuates dramatically and DTEL generally obtained the best accuracy on each data chunk.

It is worth noting that the accuracy on each chunk of CTRPrediction data may reveal an interesting rule in click habit from the website visitors. The 20,000 examples for each day are randomly extracted from the website, and each day's data are divided into two consecutive chunks in CTRPrediction. Hence, the consecutive chunks roughly embed the click habits from the first half and second half of a day, which roughly represent the working time and leisure time, respectively. From the result shown in Fig. \ref{realdata} ($d$), it can be observed that the classification performance is distinctly different on two consecutive chunks, with a close to 100\% accuracy followed by a roughly 50\% one. This indicates that the click habit changes in different time spans and also reveals the difficulty in learning the dynamic real-world data.

Table \ref{table_accuracy_real} presents the average accuracy obtained from the real-world data streams. DTEL shows a great advantage over other algorithms, which is consistent with the results obtained on the synthetic data streams. Specifically, DTEL demonstrates significantly best accuracy results on the real-world data steams, except for Electricity. AUE2 obtained the highest accuracy on the Electricity data stream. However, the value of the corresponding standard variance indicates that AUE2 is more sensitive to the concept drift in Electricity than DTEL and Learn\textsuperscript{++}.NSE. Moreover, there is no statistically significant difference among the performance of the compared algorithms on the Electricity data stream.

To make a comprehensive comparison, a Friedman test \cite{friedman} is conducted based on the average accuracy results on both synthetic data streams (Table \ref{table_accuracy_syn}) and real-world data streams (Table \ref{table_accuracy_real}), as shown in Table \ref{table_friedman}. The rank values of DTEL is 1.2895 and significantly outperform others. The Friedman test result demonstrates that the proposed DTEL is significantly better than all of the compared approaches with regard to the average accuracy value.

\begin{table}[!t]
\renewcommand{\arraystretch}{1}
\caption{Friedman Test (Nemenyi test, $CD=1.3993$ at $\alpha = 0.05$) Result Considering the Average Accuracy on Both Synthetic and Real-world Data Streams}
\centering
\setlength{\tabcolsep}{5pt}
\begin{tabular}{c c c c c c}
\hline
algorithm & DTEL & SEA & Learn\textsuperscript{++}.NSE & AUE2 & TIX \\
\hline
rank & 1.3421 & 4.4211 & 3.2632 & 2.8158 & 3.1579 \\
\hline
\end{tabular}
\label{table_friedman}
\end{table}

The runtimes of the approaches are also compared under the same computing environment (2 CPUs of 2.4 GHz Intel Core i5, 8GB main memory) in the experiment. The time complexity of DTEL at each learning step is determined by 3 factors, i.e., the chunk size $n$, the data dimensionality $d$, and the archive size of the historical model set $m$. For DTEL impelmented with decision trees, the time complexity for building a base model is $O(d^2n)$ \cite{timeTree}. In the transfer operation, the new chunk of data is first placed into the leaf nodes with $O(n)$ time complexity, where the coefficient is the height of the transferred tree. Then, an update is conducted with the new chunk of data with a time complexity of roughly $O(d^2n)$. Since DTEL needs to transfer every maintained historical model with the new chunk of data, the time efficiency of DTEL is the worst among the compared approaches. The runtime results are shown in Table \ref{table_time}. It can be observed that DTEL takes about an order of magnitude longer time than the compared algorithms in some cases. The time consumed by the transfer operations in DTEL is also presented in Table \ref{table_time}, which empirically indicates that the transfer operation is time-consuming. Since all the historical models are transferred independently in DTEL, the transfer operations can be implemented in a parallel processing manner. By parallelizing the transfer operations, the speed-up ratio is about $m$ and the runtime of DTEL could be reduced by an order of magnitude to reach a satisfactory runtime level.

\begin{table}[!t]
\renewcommand{\arraystretch}{1}
\caption{Runtime of Each Algorithm (Unit: Seconds). The Value in the Brackets after the Runtime of DTEL is Time Cost of the Transfer Operations. LNSE Represents Learn\textsuperscript{++}.NSE.}
\label{table_time}
\centering
\begin{tabular}{c c c c c c c}
\hline
Data & DTEL & SEA & LNSE & AUE2 & TIX\\
\hline
SEA200A & 2.66e2 (2.43e2) & 3.17e1 & 2.09e1 & 5.09e0 & 2.06e1 \\
SEA200G & 2.71e2 (2.48e2) & 3.20e1 & 1.97e1 & 5.73e0 & 2.04e1 \\
SEA500G & 6.28e2 (5.77e2) & 7.27e1 & 2.64e1 & 1.21e1 & 2.52e1 \\
ROT200A & 2.97e2 (2.74e2) & 9.00e1 & 8.41e1 & 6.51e0 & 3.14e1\\
ROT200G & 3.01e2 (2.79e2) & 1.03e2 & 8.40e1 & 6.70e0 & 3.13e1\\
ROT500G & 7.10e2 (6.59e2) & 2.63e2 & 1.03e2 & 1.53e1 & 3.69e1\\
CIR200A & 2.73e2 (2.51e2) & 3.42e1 & 1.99e1 & 4.91e0 & 2.73e1\\
CIR200G & 2.82e2 (2.59e2) & 4.14e1 & 2.40e1 & 4.83e0 & 3.12e1\\
CIR500G & 6.25e2 (5.74e2) & 9.68e1 & 2.94e1 & 1.25e1 & 3.38e1\\
SIN200A & 2.96e2 (2.74e2) & 3.83e1 & 2.26e1 & 4.93e0 & 2.43e1\\
SIN200G & 3.04e2 (2.81e2) & 3.90e1 & 2.19e1 & 4.90e0 & 2.28e1\\
SIN500G & 6.67e2 (6.26e2) & 9.59e1 & 2.91e1 & 1.07e1 & 2.80e1\\
STA200A & 2.14e2 (1.91e2) & 3.84e1 & 2.89e1 & 5.88e0 & 2.14e1\\
STA200G & 2.07e2 (1.85e2) & 5.04e1 & 2.10e1 & 5.99e0 & 2.12e1\\
STA500G & 7.09e2 (6.62e2) & 9.32e1 & 2.57e1 & 1.68e1 & 2.36e1\\
Covertype & 2.76e3 (2.22e3) & 1.55e3 & 1.02e3 & 5.41e2 & 9.66e2\\
PokerHand & 1.27e4 (1.18e4) & 8.83e2 & 7.75e3 & 2.86e2 & 6.09e3\\
Electricity & 1.18e2 (9.02e1) & 2.83e1 & 1.53e1 & 1.22e1 & 5.49e0\\
CTRPrediction & 3.83e2 (9.02e1) & 3.03e2 & 1.69e2 & 2.36e2 & 2.47e2\\
\hline
\end{tabular}
\end{table}

\subsection{Influence of Archive Size}
\begin{figure}[!t]
\centering
\includegraphics[width=2.3in]{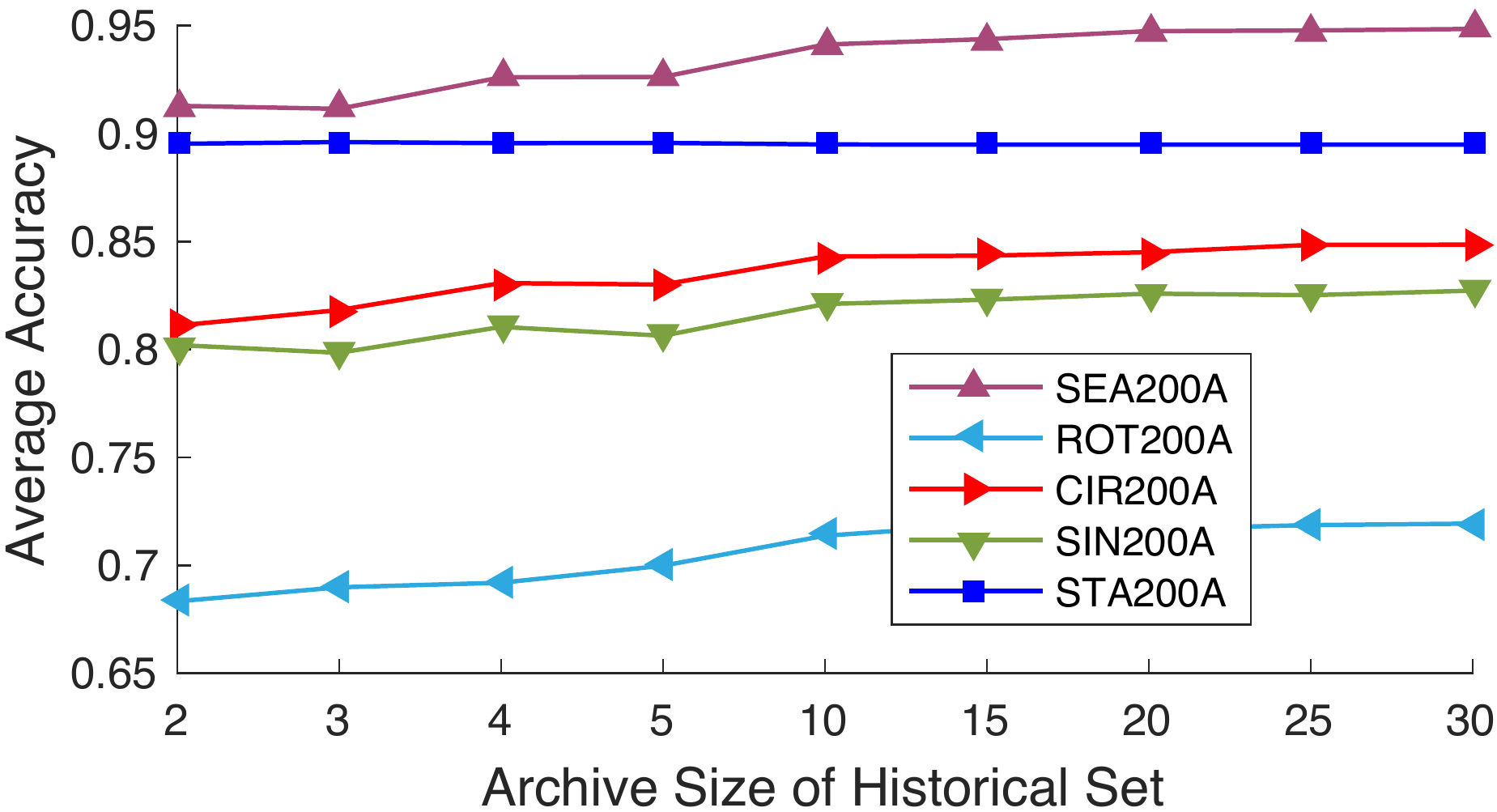}
\caption{Sensitive analysis of the size of the historical model set in DTEL.}
\label{sensitive}
\end{figure}

The influence of the only parameter in DTEL, i.e., the archive size of the historical model set $m$, is studied. The appropriate size of the historical model set may be influenced by the data distribution and the types of concept drift in the whole incremental learning process. Five data streams (i.e., SEA200A, ROT200A, CIR200A, SIN200A, and STA200A) are used to test different sizes, and the test result is shown in Fig. \ref{sensitive}. The data STA200A is generated from decision rules and the STA concept drift does not change the tree structure of each base model. Hence, a very small archive size is enough to facilitate the incremental learning with concept drift. Comprehensively considering the test result, when the archive size is smaller than 20, the average accuracy improves with the size increasing. Then, the accuracy results remain roughly stable when the value is larger. Hence, for practical applications, the archive size with a value bigger than 20 is recommended.

\section{Conclusions and Future Work}
This paper presents a new ensemble learning approach, namely DTEL, for incremental learning with concept drift. DTEL employs a diversity-based selection criterion to preserve previously trained models. Instead of being applied directly to form an ensemble for the current concept, the preserved models are further adapted to the current concept through transfer learning. Empirical studies on both synthetic and real-world data streams demonstrate the advantages of DTEL over a number of state-of-the-art incremental learning methods. 

The main potential drawback of DTEL is that it is computationally more costly than the compared methods. Although this disadvantage could be alleviated by parallel implementation of DTEL since it can be naturally parallelized, it is still worth investigating other methods to reduce the complexity of DTEL. For example, it might be unnecessary for all the preserved models to be further trained. Alternatively, some heuristic rules can be designed to identify the preserved models that is most worthy of further training. Besides, this paper only considers decision tree as the base learners of DTEL. Other base learners should be investigated. Although the general framework of DTEL (as in Algorithm 1) is not restricted to decision tree, specific transfer learning (further training) methods need to be designed for different base learners. This would also be an interesting direction for research in the future.

\section*{Acknowledgment}
The authors would like to thank...

\ifCLASSOPTIONcaptionsoff
  \newpage
\fi

\bibliographystyle{IEEEtran}
\bibliography{IEEEabrv,bare_jrnl}

\vspace{-20 mm}\begin{IEEEbiography}[]{Yu Sun}
received the B.Eng. degree in Software Engineering from Dalian University of Technology (DLUT), Dalian, Liaoning, China, in 2010, and M.Eng. degree in Software Engineering from University of Science and Technology of China (USTC), Hefei, Anhui, China, in 2013. He is currently working towards the Ph.D. degree in Computer Science with the USTC-Birmingham Joint Research Institute in Intelligent Computation and Its Applications (UBRI), School of Computer Science and Technology, USTC.

His current research concerns incremental learning and data stream mining.
\end{IEEEbiography}

\begin{IEEEbiography}[]{Ke Tang}
received the B.Eng. degree from Huazhong University of Science and Technology, Wuhan, China, in 2002, and the Ph.D. degree from Nanyang Technological University, Singapore, in 2007, respectively. 

Since 2007, he has been with the School of Computer Science and Technology, University of Science and Technology of China, where he is currently a Professor. He has authored/co-authored more than 100 refereed publications. His major research interests include evolutionary computation, machine learning, and their real-world applications.

Dr. Tang is an Associate Editor of the IEEE Transactions on Evolutionary Computation, IEEE Computational Intelligence Magazine and Computational Optimization and Applications (Springer), and served as a member of Editorial Boards for a few other journals. He is a member of the IEEE Computational Intelligence Society (CIS) Evolutionary Computation Technical Committee and the IEEE CIS Emergent Technologies Technical Committee. He is the recipient of the Royal Society Newton Advanced Fellowship.
\end{IEEEbiography}

\vspace{-50 mm}\begin{IEEEbiography}[]{Zexuan Zhu}
received the B.S. degree in computer science and technology from Fudan University, Shanghai, China, in 2003, and the Ph.D. degree in computer engineering from Nanyang Technological University, Singapore, in 2008.

He is currently an Associate Professor with the College of Computer Science and Software Engineering, Shenzhen University, Shenzhen, China. His current research interests include computational intelligence, machine learning, and bioinformatics.
\end{IEEEbiography}

\vspace{-50 mm}\begin{IEEEbiography}[]{Xin Yao}
is a Professor of Computer Science and the Director of CERCIA (the Centre of Excellence for Research in Computational Intelligence and Applications) at the University of Birmingham, UK. He is an IEEE Fellow and a Distinguished Lecturer of IEEE Computational Intelligence Society (CIS). He was the President (2014-15) of IEEE CIS. His major research interests include evolutionary computation and ensemble learning, especially online learning and class imbalance learning. His work won the 2001 IEEE Donald G. Fink Prize Paper Award, 2010 and 2015 IEEE Transactions on Evolutionary Computation Outstanding Paper Awards, 2010 BT Gordon Radley Award for Best Author of Innovation (Finalist), 2011 IEEE Transactions on Neural Networks Outstanding Paper Award, and many other best paper awards. He received the prestigious Royal Society Wolfson Research Merit Award in 2012 and the IEEE CIS Evolutionary Computation Pioneer Award in 2013. He was the Editor-in-Chief (2003-08) of IEEE Transactions on Evolutionary Computation.
\end{IEEEbiography}

\end{document}